# A Data-Driven Approach for Discovering Stochastic Dynamical Systems with Non-Gaussian Lévy Noise

Yang Li[1,2,a] and Jinqiao Duan[2,b]

[1]State Key Laboratory of Mechanics and Control of Mechanical Structures, College of Aerospace Engineering, Nanjing University of Aeronautics and Astronautics, 29 Yudao Street, Nanjing 210016, China

[2]Department of Applied Mathematics, College of Computing, Illinois Institute of Technology, Chicago, Illinois 60616, USA

[a]li_yang@nuaa.edu.cn

[b]Corresponding author: duan@iit.edu

**Abstract** With the rapid increase of valuable observational, experimental and simulating data for complex systems, great efforts are being devoted to discovering governing laws underlying the evolution of these systems. However, the existing techniques are limited to extract governing laws from data as either deterministic differential equations or stochastic differential equations with Gaussian noise. In the present work, we develop a new data-driven approach to extract stochastic dynamical systems with non-Gaussian symmetric Lévy noise, as well as Gaussian noise. First, we establish a feasible theoretical framework, by expressing the drift coefficient, diffusion coefficient and jump measure (i.e., anomalous diffusion) for the underlying stochastic dynamical system in terms of sample paths data. We then design a numerical algorithm to compute the drift, diffusion coefficient and jump measure, and thus extract a governing stochastic differential equation with Gaussian and non-Gaussian noise. Finally, we demonstrate the efficacy and accuracy of our approach by applying to several prototypical one-, two- and three-dimensional systems. This new approach will become a tool in discovering governing dynamical laws from noisy data sets, from observing or simulating complex phenomena, such as rare events triggered by random fluctuations with heavy as well as light tail statistical features.

**Keywords** Machine learning; data-driven approach; non-Gaussian Lévy noise; stochastic dynamical systems; heavy-tailed fluctuations; rare events

## 1. Introduction

Environmental and intrinsic noisy fluctuations are inevitable in complex systems in biology, physics, chemistry, engineering and finance [1–6]. The governing laws for the dynamical evolution of these systems are usually in terms of stochastic differential equations. However, the explicit analytical governing laws are often unavailable, due to lack of scientific understanding of mechanisms underlying the complex phenomena, especially in biomedical systems. Fortunately, more and more data sets for complex systems are available with the progress of scientific tools and simulation capabilities. Therefore, extracting governing laws from noisy data plays a crucial role in various scientific fields.

Data-driven techniques are recently being proposed to learn governing equations of nonlinear phenomena from data. For instance, the Sparse Identification of Nonlinear Dynamics approach was designed to discover deterministic ordinary [7] or partial [8–11] differential equations from available data sets. Boninsegna et al. [12] extended this method to learn stochastic differential equations, using Kramers-Moyal expansions. Based on the theory of Koopman generator, Klus et al. [13] generalized the Extended Dynamic Mode Decomposition method [14] to identify stochastic differential equations from data. Wu et al. [15] further developed a method to discover mean residence time and escape probability from data for stochastic differential equations. These methods are only applicable to extract either deterministic differential equations or stochastic differential equations with Gaussian noise (i.e. in terms of Brownian motion).

However, a broad range of phenomena with bursting, flights, intermittent, hopping, or rare transitions features are observed in, for example, climate change [16], gene regulation [17], ecology [18], and geophysical turbulence [19]. It is appropriate to model these phenomena via stochastic differential equations with non-Gaussian Lévy noise (i.e. in terms of Lévy motion). Stochastic differential equations with Brownian motion and Lévy motion are known to model a large and significant class of Markov processes, so called Feller processes [20,21]. Indeed, some researchers and we have recently taken these stochastic differential equations as phenomenological models for various random phenomena. Cai et al. investigated the escaping phenomena of a neuron system driven by the non-Gaussian $\alpha$-stable Lévy noise to detect its excitation behaviors [22,23]. Together with co-authors, we analyzed the transitions between the vegetative and the competence regions in a gene network under Lévy noise by quantitatively computing the exit time and maximal likely transition trajectory [6,24,25]. Based on the Greenland ice core measurement data, Ditlevsen found that the climate change system could be modeled as stochastic differential equations with Brownian motion and Lévy process [16]. Subsequently Zheng et al. developed a probabilistic framework to study the maximum likelihood climate change for an energy balance system under the combined influence of greenhouse effect and non-Gaussian $\alpha$-stable Lévy motions [1]. Some researchers found that the Lévy noise could induce stochastic resonance phenomenon in spiking neuron models [26,27]. Additionally, Liu et al. [28] studied the persistence and extinction of a delayed vaccinated SIR epidemic model with temporary immunity and Lévy jumps. Guarcello et al. [29] numerically investigated the dynamics of the current-biased long Josephson junctions by modeling it as a sine-Gordon equation driven by an oscillating field and subject to an external non-Gaussian noise.

These works show that the stochastic differential equations with Lévy motion as well as Brownian motion are insightful phenomenological models in science. Hence, it is desirable to extract such stochastic governing laws that are consistent with observational or simulated data. Based on the theory of Koopman generator, Lu and Duan [30] developed a data-driven method to learn stochastic dynamical system with Lévy noise from data in the previous work. However, this technique has a limitation that it can only estimate the Lévy noise intensity while fix the Lévy jump measure in advance. Therefore, a more efficient method needs to be proposed.

In the present work, we develop a framework to extract stochastic governing equations, with Lévy motion and Brownian motion, from noisy data. Our major contributions are: (i) to devise an effective method to extract the jump measure for the associated Lévy fluctuations, which are responsible for bursting, flights or intermittent non-Gaussian dynamical features (also related to anomalous diffusion); and (ii) to design an algorithm to compute the drift (responsible for nonlinear interactions) and the diffusion coefficient (magnitude for Gaussian fluctuations). In this new data-driven approach in discovering stochastic dynamical systems with Gaussian Brownian motion and non-Gaussian Lévy motion from data, we derive formulas to express the jump measure, drift, and diffusion coefficient, in terms of the transition probability density, inspired by the derivation of the differential Chapman-Kolmogorov equation in [31]. As the transition probability density satisfies the associated nonlocal Fokker-Planck equation, we further reformulate the jump measure, drift, and diffusion coefficient as computable quantities, via sample path data for a system under consideration. After laying this theoretical foundation, we then design numerical schemes to identify the jump measure, drift and diffusion coefficient. Numerical experiments show that our method is effective and sufficiently accurate for several prototypical examples in one-, two- and three-dimensions.

This article is organized as follows. First, we prove a theorem and a corollary to establish the theoretical framework of our approach (Section 2). That is, we derive formulas to express the jump measure, drift, and diffusion coefficient, in terms of either the transition probability density or sample paths. Then we design numerical schemes to compute the jump measure, drift and diffusion coefficient (Section 3). We conduct numerical experiments to illustrate our approach (Section 4), and finally conclude with Discussion (Section 5).

## 2. Theory and Method

Consider an $n$-dimensional stochastic dynamical system

$$\mathrm{d}\mathbf{x}(t) = \mathbf{b}(\mathbf{x}(t))\mathrm{d}t + \Lambda(\mathbf{x}(t))\mathrm{d}\mathbf{B}_t + \sigma \mathrm{d}\mathbf{L}_t, \tag{1}$$

where $\mathbf{b}(\mathbf{x}) = [b_1(\mathbf{x}), \cdots, b_n(\mathbf{x})]^T$ is the drift coefficient in $\mathbb{R}^n$, $\mathbf{B}_t = [B_{1,t}, \cdots, B_{n,t}]^T$ is $n$-dimensional Brownian motion, $\Lambda(\mathbf{x})$ is an $n \times n$ matrix and $\mathbf{L}_t = [L_{1,t}, \cdots, L_{n,t}]^T$ is symmetric Lévy motion described in the Appendix A with positive constant noise intensity $\sigma$. Assume that the initial condition is $\mathbf{x}(0) = \mathbf{z}$, $a(\mathbf{x}) = \Lambda \Lambda^T$ is the diffusion matrix, and the jump measure of $\mathbf{L}_t$ satisfies $\nu(\mathrm{d}\mathbf{y}) = W(\mathbf{y})\mathrm{d}\mathbf{y}$ for $\mathbf{y} \in \mathbb{R}^n \setminus \{0\}$. Due to the symmetry of the Lévy motion, $W(-\mathbf{y}) = W(\mathbf{y})$. For example, $W(\mathbf{y}) = c(n,\alpha)|\mathbf{y}|^{-(n+\alpha)}$ in the case of rotationally symmetric $\alpha$-stable Lévy process [21,32] with $c(n,\alpha) = \dfrac{\alpha \Gamma((n+\alpha)/2)}{2^{1-\alpha}\pi^{n/2}\Gamma(1-\alpha/2)}$, where $|\cdot|$ is the usual Euclidean norm and the Gamma function $\Gamma$ is defined as $\Gamma(z) \triangleq \int_0^\infty t^{z-1}\mathrm{e}^{-t}\mathrm{d}t$.

According to Ref. [21,33], the Fokker-Planck equation for the probability density function $p(\mathbf{x},t \mid \mathbf{z},0)$ for the solution of Eq. (1) is then

$$\begin{aligned}\frac{\partial p}{\partial t} = &-\nabla \cdot \left[\mathbf{b}p(\mathbf{x},t \mid \mathbf{z},0)\right] + \frac{1}{2}Tr\left[H\left(ap(\mathbf{x},t \mid \mathbf{z},0)\right)\right] \\ &+ \int_{\mathbb{R}^n \setminus \{0\}} \left[p(\mathbf{x}+\sigma \mathbf{y},t \mid \mathbf{z},0) - p(\mathbf{x},t \mid \mathbf{z},0)\right]W(\mathbf{y})\mathrm{d}\mathbf{y},\end{aligned} \tag{2}$$

where the symbol $H$ denotes Hessian matrix $\left(\partial_{x_i x_j}\right)$ and we interpret $H(ap)$ as matrix multiplications of $H$ and $ap$ (note that $p$ is scalar). The integral in the right hand side is understood as a Cauchy principal value integral and the initial condition is $p(\mathbf{x},0|\mathbf{z},0) = \delta(\mathbf{x}-\mathbf{z})$.

In order to discover stochastic dynamical systems with non-Gaussian Lévy noise from data, we derive the following theorem. It expresses jump measure, drift and diffusion in terms of solution of Fokker-Planck equation.

**Theorem 1**. *For every $\varepsilon > 0$, the probability density function $p(\mathbf{x},t|\mathbf{z},0)$ and the jump measure, drift and diffusion have the following relations:*

i) *For every $\mathbf{x}$ and $\mathbf{z}$ satisfying $|\mathbf{x}-\mathbf{z}| > \varepsilon$,*

$$\lim_{t \to 0} p(\mathbf{x},t|\mathbf{z},0)/t = \sigma^{-n} W(\sigma^{-1}(\mathbf{x}-\mathbf{z}))$$

*uniformly in $\mathbf{x}$ and $\mathbf{z}$;*

ii) *For $i = 1, 2, \ldots, n$,*

$$\lim_{t \to 0} t^{-1} \int_{|\mathbf{x}-\mathbf{z}|<\varepsilon} (x_i - z_i) p(\mathbf{x},t|\mathbf{z},0) d\mathbf{x} = b_i(\mathbf{z});$$

iii) *For $i, j = 1, 2, \ldots, n$,*

$$\lim_{t \to 0} t^{-1} \int_{|\mathbf{x}-\mathbf{z}|<\varepsilon} (x_i - z_i)(x_j - z_j) p(\mathbf{x},t|\mathbf{z},0) d\mathbf{x} = a_{ij}(\mathbf{z}) + \sigma^{-n} \int_{|\mathbf{y}|<\varepsilon} y_i y_j W(\sigma^{-1}\mathbf{y}) d\mathbf{y}.$$

Remark that for the implementation of subsequent numerical algorithm, we reformulate these relations from Theorem 1 in the following corollary. Thus, the jump measure, drift and diffusion are in terms of the sample paths of the stochastic differential equation (1).

**Corollary 2**. *For every $\varepsilon > 0$, the solution $\mathbf{x}(t)$ of the stochastic differential equation (1) and the jump measure, drift and diffusion have the following relations:*

i) *For every $m > 1$,*

$$\lim_{t \to 0} t^{-1} \mathbb{P}\{|\mathbf{x}(t) - \mathbf{z}| \in [\varepsilon, m\varepsilon] | \mathbf{x}(0) = \mathbf{z}\} = \sigma^{-n} \int_{|\mathbf{y}| \in [\varepsilon, m\varepsilon)} W(\sigma^{-1}\mathbf{y}) d\mathbf{y};$$

ii) *For $i = 1, 2, \ldots, n$,*

$$\lim_{t \to 0} t^{-1} \mathbb{P}\{|\mathbf{x}(t) - \mathbf{z}| < \varepsilon | \mathbf{x}(0) = \mathbf{z}\} \cdot \mathrm{E}\left[(x_i(t) - z_i) | \mathbf{x}(0) = \mathbf{z}; |\mathbf{x}(t) - \mathbf{z}| < \varepsilon \right] = b_i(\mathbf{z});$$

iii) *For $i, j = 1, 2, \ldots, n$.*

$$\lim_{t \to 0} t^{-1} \mathbb{P}\{|\mathbf{x}(t) - \mathbf{z}| < \varepsilon | \mathbf{x}(0) = \mathbf{z}\} \cdot \mathrm{E}\left[(x_i(t) - z_i)(x_j(t) - z_j) | \mathbf{x}(0) = \mathbf{z}; |\mathbf{x}(t) - \mathbf{z}| < \varepsilon \right]$$
$$= a_{ij}(\mathbf{z}) + \sigma^{-n} \int_{|\mathbf{y}|<\varepsilon} y_i y_j W(\sigma^{-1}\mathbf{y}) d\mathbf{y}.$$

The proofs of Theorem 1 and Corollary 2 are in Appendix D and E.

## 3. Numerical algorithms

In this section, we will propose a numerical approach to extract a stochastic dynamical system with a drift coefficient, a diffusion coefficient and an additive Lévy noise in the form of Eq. (1) based on the Corollary 2 in Section 2. For the sake of concreteness, we take the rotationally symmetric $\alpha$-stable Lévy motion as an example to show the effectiveness of our method. In fact, the $\alpha$-stable case is more difficult than the others such as exponentially light jump process [34] since its kernel function $W(\mathbf{y})$ of the jump measure approaches infinity as $\mathbf{y} \to 0$. Therefore, this method can be naturally generalized to the other cases if it can work on the $\alpha$-stable case.

Assume that there exists a pair of data sets containing $M$ elements, respectively,

$$
\begin{aligned}
Z &= [\mathbf{z}_1, \mathbf{z}_2, \cdots, \mathbf{z}_M], \\
X &= [\mathbf{x}_1, \mathbf{x}_2, \cdots, \mathbf{x}_M],
\end{aligned}
\tag{3}
$$

where $\mathbf{x}_j$ is the image of $\mathbf{z}_j$ after a small evolution time $h$. In addition, we also need to choose a dictionary of basis functions $\Psi(\mathbf{x}) = [\psi_1(\mathbf{x}), \psi_2(\mathbf{x}), \cdots, \psi_K(\mathbf{x})]$. On the basis of the data sets and basis functions, what we want to identify are the kernel function and noise intensity of the Lévy motion, the drift coefficient and the diffusion matrix.

### 3.1. Algorithm for identification of the Lévy motion

For the sake of simplicity, we first consider the one-dimensional case. Based on the first assertion of Corollary 2, it is seen that the two sides of the equation only depend on the distance of $x$ and $z$ instead of their specific positions. Thus we construct a new data set $R = [|y_1|, |y_2|, \cdots, |y_M|]$ with $y_j = x_j - z_j$. Therefore, the probability in the left hand side of the first assertion can be approximated by the ratio of the number of the points falling into the interval $[\varepsilon, m\varepsilon)$ to the total number $M$.

In order to numerically identify the stability parameter $\alpha$ and the noise intensity $\sigma$, we consider the integration on $N+1$ intervals $[\varepsilon, m\varepsilon), [m\varepsilon, m^2\varepsilon), \ldots, [m^N\varepsilon, m^{N+1}\varepsilon)$ with the positive integer $N$, the positive real number $\varepsilon$ and the real number $m > 1$. Assume that there are $n_0, n_1, \ldots, n_N$ points from the data set $R$ which fall into these intervals respectively. Therefore,

$$h^{-1}\mathbb{P}\left\{|x(h) - z| \in [m^k\varepsilon, m^{k+1}\varepsilon) \Big| x(0) = z\right\} \approx h^{-1}M^{-1}n_k.$$

On the other hand, the integration from the right hand side yields

$$\sigma^{-1}\int_{|y|\in[m^k\varepsilon, m^{k+1}\varepsilon)} W(\sigma^{-1}y)dy$$
$$= \sigma^{\alpha} c(1,\alpha)\int_{|y|\in[m^k\varepsilon, m^{k+1}\varepsilon]} |y|^{-(1+\alpha)}dy$$
$$= 2\sigma^{\alpha} c(1,\alpha)\alpha^{-1}\varepsilon^{-\alpha} m^{-k\alpha}(1-m^{-\alpha}).$$

Combining the two equations, we have $N+1$ equalities

$$2\sigma^{\alpha} c(1,\alpha)\alpha^{-1}\varepsilon^{-\alpha} m^{-k\alpha}(1-m^{-\alpha}) = h^{-1}M^{-1}n_k, \quad k = 0, 1, \ldots, N. \tag{4}$$

The ratios of the first equation to the other $N$ equations lead to the solutions

$$\alpha = (k\ln m)^{-1}\ln\frac{n_0}{n_k}, \quad k = 1, 2, \ldots, N. \tag{5}$$

Denote $\tilde{\alpha}$, $\tilde{\sigma}$ as the numerical approximation of the parameters $\alpha$, $\sigma$, respectively. If $N=1$, Eq. (5) is the expression of $\tilde{\alpha}$. In order to make full use of the data information, we can choose bigger $N$ and identify the estimate value $\tilde{\alpha}$ as the Arithmetic Mean of Eqs. (5). Then in terms of Eqs. (4), the noise intensity is calculated as

$$\sigma = \left[\frac{\tilde{\alpha}\varepsilon^{\tilde{\alpha}} m^{k\tilde{\alpha}} n_k}{2c(1,\tilde{\alpha})hM(1-m^{-\tilde{\alpha}})}\right]^{1/\tilde{\alpha}}, \quad k = 0, 1, \ldots, N. \tag{6}$$

Hence, the approximate noise intensity $\tilde{\sigma}$ can be identified as the Arithmetic Mean of Eqs. (6).

The multi-dimensional cases of rotationally symmetric $\alpha$-stable Lévy noise are similar yet with some changes. Taking two-dimensional system for example, we still consider $N+1$ intervals $[\varepsilon, m\varepsilon), [m\varepsilon, m^2\varepsilon), \ldots, [m^N\varepsilon, m^{N+1}\varepsilon)$. Assume that there are $n_0, n_1, \ldots, n_N$ points from the data set $R = [|\mathbf{y}_1|, |\mathbf{y}_2|, \cdots, |\mathbf{y}_M|]$ with $\mathbf{y}_j = \mathbf{x}_j - \mathbf{z}_j$ which fall into these intervals respectively. The probability on the annular region is equal to

$$h^{-1}\mathbb{P}\left\{|\mathbf{x}(t)-\mathbf{z}|\in[m^k\varepsilon, m^{k+1}\varepsilon)\Big|\mathbf{x}(0)=\mathbf{z}\right\} \approx h^{-1}M^{-1}n_k.$$

On the other hand, the application of polar transformation on another side leads to

$$\sigma^{-2}\int_{|\mathbf{y}|\in[m^k\varepsilon, m^{k+1}\varepsilon)} W(\sigma^{-1}\mathbf{y})d\mathbf{y}$$
$$= \sigma^{\alpha} c(2,\alpha)\int_{|\mathbf{y}|\in[m^k\varepsilon, m^{k+1}\varepsilon]} |\mathbf{y}|^{-(2+\alpha)}d\mathbf{y}$$
$$= \sigma^{\alpha} c(2,\alpha)\int_0^{2\pi}d\theta\int_{m^k\varepsilon}^{m^{k+1}\varepsilon} r^{-(1+\alpha)}dr$$
$$= 2\pi\sigma^{\alpha} c(2,\alpha)\alpha^{-1}\varepsilon^{-\alpha} m^{-k\alpha}(1-m^{-\alpha}).$$

Combining these equations, we obtain the approximate parameter

$$\alpha = (k \ln m)^{-1} \ln \frac{n_0}{n_k}, \quad k = 1, 2, \ldots, N \tag{7}$$

and the estimate noise intensity

$$\sigma = \left[ \frac{\tilde{\alpha}\varepsilon^{\tilde{\alpha}} m^{k\tilde{\alpha}} n_k}{2\pi c(2,\tilde{\alpha}) hM \left(1 - m^{-\tilde{\alpha}}\right)} \right]^{1/\tilde{\alpha}}, \quad k = 0, 1, \ldots, N. \tag{8}$$

Consequently, the expressions of the approximate stability parameter and the noise intensity can be naturally derived for arbitrary dimensional system in terms of the Jacobian determinant of the polar coordinates transformation. For example, the Jacobian determinant is $r^2 \sin\varphi$ for the three-dimensional system. Therefore, the approximate stability parameter is deduced as

$$\alpha = (k \ln m)^{-1} \ln \frac{n_0}{n_k}, \quad k = 1, 2, \ldots, N \tag{9}$$

and the noise intensity is

$$\sigma = \left[ \frac{\tilde{\alpha}\varepsilon^{\tilde{\alpha}} m^{k\tilde{\alpha}} n_k}{4\pi c(3,\tilde{\alpha}) hM \left(1 - m^{-\tilde{\alpha}}\right)} \right]^{1/\tilde{\alpha}}, \quad k = 0, 1, \ldots, N. \tag{10}$$

It is important to emphasize that this method avoids the problem of curse of dimensionality since the data information for arbitrary dimension system is transformed into positive real number set $R$. In other words, the increase of dimensionality does not necessarily require more data.

### 3.2. Algorithm for identification of the drift term

Additionally, we need to identify the drift coefficient $\mathbf{b}(\mathbf{x})$ in terms of the dictionary of basis functions $\Psi(\mathbf{x})$. Every component of the drift coefficient can be approximated as $b_i(\mathbf{x}) \approx \sum_{k=1}^{K} c_{i,k} \psi_k(\mathbf{x})$, $i = 1, 2, \ldots, n$. According to the second assertion of Corollary 2, the computation of the drift term just requires the data satisfying $|\mathbf{x} - \mathbf{z}| < \varepsilon$. Therefore, we record the positions of the elements smaller than $\varepsilon$ in the data set $R$ and delete the others. After deleting the data of the corresponding positions in $Z$ and $X$, we obtain the new data sets with $\hat{M}$ elements respectively

$$\begin{aligned} \hat{Z} &= \left[ \hat{\mathbf{z}}_1, \hat{\mathbf{z}}_2, \cdots, \hat{\mathbf{z}}_{\hat{M}} \right], \\ \hat{X} &= \left[ \hat{\mathbf{x}}_1, \hat{\mathbf{x}}_2, \cdots, \hat{\mathbf{x}}_{\hat{M}} \right]. \end{aligned} \tag{11}$$

Since the drift and diffusion terms only contribute to the continuous portion, the probability of $|\mathbf{x}-\mathbf{z}|>\varepsilon$ mainly comes from occasional big jumps of Lévy noise. Thus the conditional probability $\mathbb{P}\{|\mathbf{x}(t)-\mathbf{z}|<\varepsilon|\mathbf{x}(0)=\mathbf{z}\}$ is less affected by the initial position $\mathbf{z}$ and it can be estimated as $\hat{M}M^{-1}$. In addition, the limit expressions in the left hand side of Corollary 2(ii) can be approximated by finite differences. Above all, we derive the following group of equations

$$A\mathbf{c}_i = B_i,$$
$$A = \begin{bmatrix} \psi_1(\hat{\mathbf{z}}_1) & \cdots & \psi_K(\hat{\mathbf{z}}_1) \\ \vdots & \ddots & \vdots \\ \psi_1(\hat{\mathbf{z}}_{\hat{M}}) & \cdots & \psi_K(\hat{\mathbf{z}}_{\hat{M}}) \end{bmatrix}, \quad (12)$$
$$B_i = \hat{M}M^{-1}h^{-1}\left[\hat{x}_{i,1} - \hat{z}_{i,1}, \cdots, \hat{x}_{i,\hat{M}} - \hat{z}_{i,\hat{M}}\right]^T.$$

Here, $\mathbf{c}_i = [c_{i,1}, c_{i,2}, \ldots, c_{i,K}]^T$. This group of equations has no solutions in general since the number of the equations is much more than the one of variables. Consequently, we turn to find the optimal solution of the least squares problem $\min_{\mathbf{c}_i}\|A\mathbf{c}_i - B_i\|_2$. It is well known that this leads to the solution

$$\tilde{\mathbf{c}}_i = (A^T A)^{-1}(A^T B_i). \quad (13)$$

### 3.3. Algorithm for identification of the diffusion term

Finally, it remains to find the diffusion term $a(\mathbf{x})$. Based on the dictionary of basis functions $\Psi(\mathbf{x})$, every component of the diffusion coefficient can be approximated as $a_{ij}(\mathbf{x}) \approx \sum_{k=1}^{K} d_{ij,k}\psi_k(\mathbf{x})$, $i,j = 1, 2, \ldots, n$. It is found that the computation of the diffusion term also requires the data satisfying $|\mathbf{x}-\mathbf{z}|<\varepsilon$, i.e., the data in the sets $\hat{Z}$ and $\hat{X}$ in Eq. (11). Based on the third assertion of Corollary 2, we derive the following group of equations

$$A\mathbf{d}_{ij} = B_{ij},$$
$$A = \begin{bmatrix} \psi_1(\hat{\mathbf{z}}_1) & \cdots & \psi_K(\hat{\mathbf{z}}_1) \\ \vdots & \ddots & \vdots \\ \psi_1(\hat{\mathbf{z}}_{\hat{M}}) & \cdots & \psi_K(\hat{\mathbf{z}}_{\hat{M}}) \end{bmatrix}, \quad (14)$$
$$B_{ij} = \hat{M}M^{-1}h^{-1}\left[(\hat{x}_{i,1}-\hat{z}_{i,1})(\hat{x}_{j,1}-\hat{z}_{j,1}), \cdots, (\hat{x}_{i,\hat{M}}-\hat{z}_{i,\hat{M}})(\hat{x}_{j,\hat{M}}-\hat{z}_{j,\hat{M}})\right]^T - S_{ij}(\varepsilon),$$
$$S_{ij}(\varepsilon) = \tilde{\sigma}^{-n}\int_{|\mathbf{y}|<\varepsilon} y_i y_j W(\tilde{\sigma}^{-1}\mathbf{y})d\mathbf{y}.$$

Here, $\mathbf{d}_{ij} = [d_{ij,1}, d_{ij,2}, \ldots, d_{ij,K}]^T$ and the computation of $S_{ij}(\varepsilon)$ requires the estimated parameters $\tilde{\alpha}$ and $\tilde{\sigma}$ in Subsection 3.1. Subsequently, we still consider the least squares problem $\min_{\mathbf{d}_{ij}} \|A\mathbf{d}_{ij} - B_{ij}\|_2$ to find the optimal solution. It is well known that this leads to the solution

$$\tilde{\mathbf{d}}_{ij} = \left(A^T A\right)^{-1} \left(A^T B_{ij}\right). \tag{15}$$

Since the diffusion matrix is symmetric, we just need to compute the coefficients for $i = 1, 2, \ldots, n$, $j = i, i+1, \ldots, n$.

## 4. Examples

In this section, we will present several examples to illustrate our approach for discovering stochastic dynamical systems from simulated data sets. Prototypical examples are chosen for one-, two- and three-dimensional cases. In what follows, the basis functions to approximate drift and diffusion terms are selected to be polynomials if not specially stated.

**Example 1** Consider the one-dimensional double-well stochastic dynamical system

$$dx(t) = \left(4x(t) - x^3(t)\right)dt + \left(1 + x(t)\right)dB_t + 2dL_t.$$

Then we have the drift coefficient $b(x) = 4x - x^3$, the diffusion coefficient $a(x) = 1 + 2x + x^2$ and the Lévy noise intensity $\sigma = 2$. The time step is fixed as $h = 0.001$ and the chosen $10^7$ initial points $Z = [z_1, z_2, \ldots, z_M]$ are distributed uniformly within the interval $[-3, 3]$. Then the data set $X$ is generated by Euler scheme with initial points from $Z$. Specifically, $x_j = z_j + b(z_j)h + h^{1/2}\Lambda(z_j)B_1 + \sigma h^{1/\alpha}L_1$, $j = 1, 2, \ldots, M$ where $B_1 \sim N(0,1)$ and $L_1 \sim S_\alpha(1,0,0)$. Here we utilize the program of Veillette [35] to simulate the $\alpha$-stable random variable $L_1$.

Next we take the three cases of $\alpha = 0.5, 1, 1.5$ into consideration. First of all, the set $R = [|y_1|, |y_2|, \cdots, |y_M|]$ is constituted by the component $y_j = x_j - z_j$. Then we choose the parameters as $N = 2$, $\varepsilon = 1$ and $m = 5$. Based on the data set $R$, Eqs. (5) and (6) are calculated to generate the values of the stability parameter $\alpha$ and the Lévy noise intensity $\sigma$ which are listed in Table 1. It is seen that all these values are perfectly consistent with the true parameters.

In order to obtain the drift and diffusion terms, we first construct the data sets $\hat{Z}$ and $\hat{X}$ according to $R$ and $\varepsilon$. Then $S(\varepsilon) = 2\tilde{\sigma}^{\tilde{\alpha}} \varepsilon^{2-\tilde{\alpha}} c(1, \tilde{\alpha})(2-\tilde{\alpha})^{-1}$. The matrices $A$, $B_i$ and $B_{ij}$ can be generated based on Section 3. Employing Eqs. (13) and (15), we compute the least square solutions $\tilde{\mathbf{c}}$ and $\tilde{\mathbf{d}}$ for the three cases $\alpha = 0.5, 1, 1.5$.

In general, the solutions of Eqs. (13) and (15) will not be sparse. To avoid the over-fitting or under-fitting regimes, we expect an excessively sparse solution containing only a small portion of the values in Eqs. (13) and (15). For this purpose, the Cross Validation procedure is used to select solutions with optimal sparsity in this paper. For its details, please refer to Ref. [12].

The sparse solutions of the drift and diffusion coefficients are computed and listed in Tables 2 and 3. It is found that the learning drift and diffusion terms are sufficiently accurate to the original ones, which affirms the effectiveness of our method.

It is necessary to discuss the criterion of the choices of so many parameters used in the estimation process. Firstly, it is found that the various values of $m$ ranging from 2 to 10 lead to no obvious changes to the results. The different selections of $N$ also have few impacts. Of course, they cannot be chosen too large since the points located at the last interval $[m^N \varepsilon, m^{N+1}\varepsilon)$ will be so rare and such circumstance will result in large error. Moreover, remark that the convergence of $p(\mathbf{x},t\,|\,\mathbf{z},0)/t$ to $\sigma^{-n}W(\sigma^{-1}(\mathbf{x}-\mathbf{z}))$ in the first assertion of Theorem 1 requires the condition $|\mathbf{x}-\mathbf{z}|>\varepsilon$. Therefore, for small but finite $h$ numerically, the choice of $\varepsilon$ should satisfy $\varepsilon \gg h$ to guarantee the accuracy. Table 4 lists the results of the estimate parameter $\tilde{\alpha}$ for several values of $\varepsilon$ and $h$. It is found that the conditions $\varepsilon > 0.5$ and $\varepsilon > 1.5$ are sufficiently accurate for the cases of $h=0.001$ and $h=0.01$ respectively. Note that larger $\varepsilon$ should be selected with the increase of $h$. Meanwhile, we can also conclude from this table that this method is robust enough for different values of the time step $h$.

**Example 2** Consider the two-dimensional Maier-Stein stochastic dynamical system

$$dx_1 = (x_1 - x_1^3 - 5x_1x_2^2)dt + (1+x_2)dB_{1,t} + dB_{2,t} + 2dL_{1,t},$$
$$dx_2 = -(1+x_1^2)x_2 dt + x_1 dB_{2,t} + 2dL_{2,t},$$

where $\mathbf{B}_t = [B_{1,t}, B_{2,t}]^T$ is a two-dimensional Brownian motion and $\mathbf{L}_t = [L_{1,t}, L_{2,t}]^T$ is a two-dimensional rotationally symmetric $\alpha$-stable Lévy motion. Then we have the drift coefficient $\mathbf{b}(\mathbf{x}) = [x_1 - x_1^3 - 5x_1x_2^2, -(1+x_1^2)x_2]^T$, the diffusion coefficient $a(\mathbf{x}) = \begin{bmatrix} 2+2x_2+x_2^2 & x_1 \\ x_1 & x_1^2 \end{bmatrix}$ and the Lévy noise intensity $\sigma = 2$. Maier-Stein system is a well-known two-dimensional double-well model in stochastic dynamical systems with complicated dynamical behaviors [36]. The time step is also fixed as $h = 0.001$ and the chosen initial points $Z = [\mathbf{z}_1, \mathbf{z}_2, ..., \mathbf{z}_M]$ are distributed uniformly within the area $[-2, 2] \times [-2, 2]$ with a mesh $10000 \times 10000$. Then the data set $X$ is generated by Euler scheme with initial points from $Z$. Specifically, $\mathbf{x}_j = \mathbf{z}_j + b(\mathbf{z}_j)h + h^{1/2}\Lambda(\mathbf{z}_j)\mathbf{B}_1 + \sigma \mathbf{L}_h$, $j = 1, 2, ..., M$, where $\mathbf{B}_1 \sim N(0, I)$.

In order to simulate rotationally symmetric stable random vector $\mathbf{L}_h$, we adopt Nolan's method [37]. Let $F \sim S_{\alpha/2}\left(2\gamma_0^2 (\cos(\pi\alpha/4))^{2/\alpha}, 1, 0\right)$ be a random variable where the scaling parameter is $2\gamma_0^2 (\cos(\pi\alpha/4))^{2/\alpha}$, the skewness parameter is 1 and the shift parameter is zero, and $\mathbf{G} \sim N(0, I)$ be two-dimensional standard normal random vector which is independent of $F$. Then the random vector $\mathbf{X} = F^{1/2}\mathbf{G}$ is rotationally symmetric with the characteristic function

$$E\exp(i\mathbf{u}^T \mathbf{X}) = \exp\left(-\gamma_0^\alpha |\mathbf{u}|^\alpha\right).$$

According to Ref. [21], the characteristic function of $\mathbf{L}_h$ is

$$E\exp(i\mathbf{u}^T \mathbf{L}_h) = \exp\left(-h|\mathbf{u}|^\alpha\right).$$

Thus let $\gamma_0 = h^{1/\alpha}$ and then we can simulate $\mathbf{L}_h$ by the random vector $\mathbf{X}$.

As before, the set $R = [r_1, r_2, ..., r_M]$ is constituted by the component $r_j = |\mathbf{x}_j - \mathbf{z}_j|$. The parameters are still fixed as $N = 2$, $\varepsilon = 1$ and $m = 5$. Based on the data set $R$, the values of the stability parameter $\alpha$ and the Lévy noise intensity $\sigma$ are estimated by Eqs. (7) and (8), which are listed in Table 5. It is seen that the results agree well with the true parameters for the three cases $\alpha = 0.5, 1, 1.5$.

On the other hand, the drift and diffusion terms can be calculated by applying the least squares method separately for their components. Therein, it is computed that $S_{11}(\varepsilon) = S_{22}(\varepsilon) = \pi\tilde{\sigma}^{\tilde{\alpha}}\varepsilon^{2-\tilde{\alpha}}c(2, \tilde{\alpha})(2-\tilde{\alpha})^{-1}$ and $S_{12}(\varepsilon) = 0$ in terms of polar transformation. The results are listed in Tables 6 and 7 and agree well with the true values.

**Example 3** Consider the three-dimensional classical Lorenz stochastic dynamical system

$$\begin{aligned}
dx_1 &= 10(-x_1 + x_2)dt + (1 + x_3)dB_{1,t} + dB_{2,t} + 2dL_{1,t}, \\
dx_2 &= (4x_1 - x_2 - x_1 x_3)dt + x_2 dB_{2,t} + 2dL_{2,t}, \\
dx_3 &= (-8/3 \, x_3 + x_1 x_2)dt + x_1 dB_{3,t} + 2dL_{3,t}.
\end{aligned}$$

Then we have the Lévy noise intensity $\sigma = 2$ and the drift and the diffusion coefficients

$$\mathbf{b}(\mathbf{x}) = \left[10(-x_1 + x_2), 4x_1 - x_2 - x_1 x_3, -8/3 x_3 + x_1 x_2\right]^T,$$

$$a(\mathbf{x}) = \begin{bmatrix} 2 + 2x_3 + x_3^2 & x_2 & 0 \\ x_2 & x_2^2 & 0 \\ 0 & 0 & x_1 \end{bmatrix}.$$

The time step is still fixed as $h = 0.001$ and the chosen initial points $Z = [\mathbf{z}_1, \mathbf{z}_2, ..., \mathbf{z}_M]$ are distributed uniformly within the area $[-2, 2] \times [-2, 2] \times [-2, 2]$ with a mesh $400 \times 400 \times 400$. Then the data set $X$ is generated by Euler scheme with initial points from $Z$. Specifically, $\mathbf{x}_j = \mathbf{z}_j + b(\mathbf{z}_j)h + h^{1/2}\Lambda(\mathbf{z}_j)\mathbf{B}_1 + \sigma \mathbf{L}_h$, $j = 1, 2, ..., M$. Therein, we still adopt Nolan's method to simulate the random vector $\mathbf{L}_h$ as Example 2.

We still take the three cases of $\alpha = 0.5, 1, 1.5$ into consideration and choose the parameters as $N = 2$, $\varepsilon = 1$ and $m = 5$. Based on the data set $R = [r_1, r_2, ..., r_M]$, we estimate the values of the stability parameter $\alpha$ and the Lévy noise intensity $\sigma$ by Eqs. (9) and (10), and list the results in Table 8. As we can see, the results agree well with the true parameters for the three cases $\alpha = 0.5, 1, 1.5$. Using polar transformation, we have $S_{ii}(\varepsilon) = 4/3\pi\tilde{\sigma}^{\tilde{\alpha}}\varepsilon^{2-\tilde{\alpha}}c(3, \tilde{\alpha})(2-\tilde{\alpha})^{-1}$ and $S_{ij}(\varepsilon) = 0$ for $i \neq j$. It is observed that the learning drift and diffusion terms in Tables 9 and 10 are sufficiently accurate to the original ones, verifying the effectiveness of our method to the three-dimensional system.

**Example 4** Consider a genetic regulatory system driven by both Gaussian Brownian noise and non-Gaussian Lévy noise, and with rational drift and diffusion coefficients [38]

$$dx(t) = \left[\frac{k_f x^2(t)}{x^2(t) + K_d} - k_d x(t) + R_{bas}\right]dt + \frac{x(t)}{\sqrt{x^2(t) + 0.5}} dB_t + 2dL_t,$$

where the system parameters are $k_f = 6\min^{-1}$, $K_d = 10$, $k_d = 1\min^{-1}$, and $R_{bas} = 0.4\min^{-1}$. Then we have the drift coefficient $b(x) = k_f x^2/(x^2 + K_d) - k_d x + R_{bas}$, the diffusion coefficient $a(x) = x^2/(x^2 + 0.5)$ and the Lévy noise intensity $\sigma = 2$. The time step is fixed as $h = 0.001$ and the $10^7$ chosen initial points $Z = [z_1, z_2, ..., z_M]$ are distributed uniformly within the interval $[0, 5]$. Then the data set $X$ is generated by Euler scheme with initial points from $Z$, analogous to Example 1. The dictionary of basis functions is chosen as

$$\Psi(x) = [1, x, x^2, x^3, \sin x, \cos 11x, \sin 11x,$$
$$-10\tanh^2(10x)+10, -10\tanh^2(10x-10)+10,$$
$$\exp\{-50x^2\}, \exp\{-50(x-3)^2\}, \exp\{-0.3x^2\},$$
$$\exp\{-0.3(x-3)^2\}, \exp\{-2(x-2)^2\}, \exp\{-50(x-4)^2\},$$
$$\exp\{-0.6(x-4)^2\}, \exp\{-0.6(x-3)^2\},$$
$$-2\tanh^2(2x-4)+2, \tanh^2(x-4)+1].$$

We still consider three cases $\alpha = 0.5, 1, 1.5$ and choose the parameters as $N=2$, $\varepsilon=1$ and $m=5$. Then the stability parameter $\alpha$ and noise intensity $\sigma$ are estimated and listed in Table 11. It is found that they are consistent with the true values. Based on Eqs. (13) and (15), we can calculate the drift and diffusion terms as

$$\tilde{\mathbf{c}} = [-230.1093, -58.4362, 39.0868, -4.2768,$$
$$44.4308, 0, 0, 0, 0, 0, 129.5482, 97.7971,$$
$$0, 11.0930, 0, 35.4622, -39.2297, 0, 47.3796]^T,$$
$$\tilde{\mathbf{d}} = [-371.0493, -111.8190, 73.3493, -7.7591,$$
$$79.7572, 0, 0, 0, 0, -0.3813, 0, 273.9994,$$
$$257.1121, -3.8454, 7.2374, 0, 20.0640,$$
$$-105.2488, 0.3325, 40.3377]^T.$$

For comparison, we plot true and learned functions of drift and diffusion in Fig. 1. It is seen that they agree well in the domain of interest $[0, 5]$. This implies that our method is also valid for non-polynomial cases and non-polynomial basis functions.

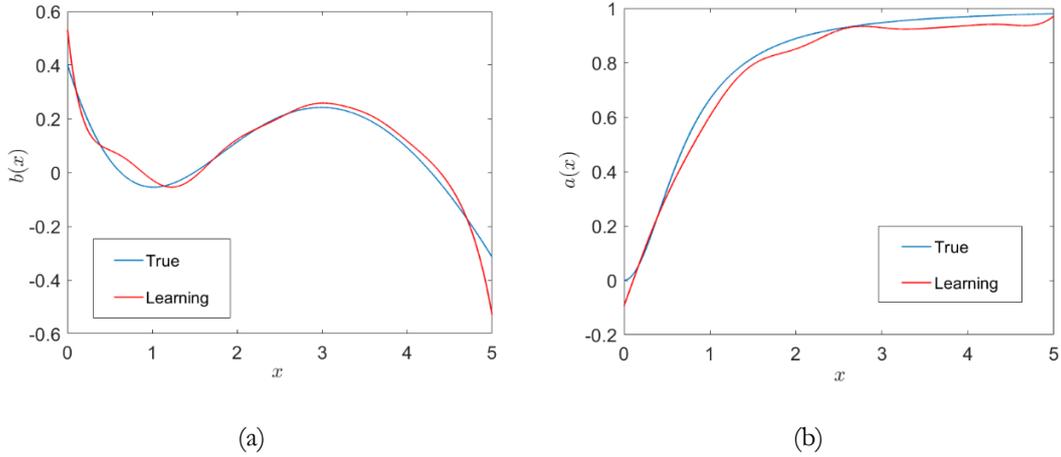

(a) (b)

Fig. 1. (a) True and learned drift coefficients. (b) True and learned diffusion coefficients

## 5. Discussion

In this paper, we have devised a new data-driven approach to extract stochastic dynamical systems with both non-Gaussian Lévy noise and Gaussian Brownian noise, from observational, experimental or simulated data sets. More specifically, we have derived formulas which express the jump measure, drift and diffusion coefficient in terms of data for sample paths. Then we have designed numerical algorithms and tested on prototypical systems. The existing works in this topic have only dealt with systems with Gaussian noise. Our new approach provides a data-driven tool to extract stochastic governing laws for complex phenomena under non-Gaussian fluctuations.

Theorem 1 may be regarded as a non-Gaussian Lévy version of the well-known Kramers-Moyal formulas [39,40], and is applicable to Markov processes with jumps, such as those modeled as the solution processes of stochastic differential equations with Brownian motion and Lévy motion. In the derivation of drift and diffusion coefficient in Theorem 1 (ii)-(iii), the domain of integration is reduced from the whole space to a small neighborhood in order to guarantee the convergence. As known in Section 2, the function $W$ is defined as the Radon-Nikodym derivative of the jump measure with respect to the Lebesgue measure. By Theorem 1 (i) and the Lindeberg condition for sample path continuity [39], we see that when $W$ is zero, the corresponding Markov process has continuous sample paths. Indeed, the function $W$ actually quantifies the jump frequency and jump size.

Our approach also applies to stochastic dynamical systems with non-symmetric Lévy motion (i.e., when $W$ is not an even function), and with other types of Lévy processes, such as tempered stable processes [41] or exponentially light jump processes [34]. It also applies to high dimensional systems driven by Lévy noise, but the numerical aspects in this case will be more complicated.

Finally, we note that it is a challenge to extend our approach to data sets from stochastic dynamical systems with multiplicative Lévy noise (i.e., the noise intensity $\sigma$ in equation (1) depending on system state $\mathbf{x}$). In this case, the multiplicative Lévy noise destroys the "space homogeneity" in the first assertion of Theorem 1, leading to the fact that the function $W$ depends on both $\mathbf{x}$ and $\mathbf{z}$ (not just depends only on the spatial translation $\mathbf{x}-\mathbf{z}$).

## Acknowledgements

We would like to thank Ting Gao, Xiaoli Chen, Min Dai, Jianyu Hu, Yanxia Zhang and Yubin Lu for helpful discussions. This research was supported by the National Natural Science Foundation of China (No. 11772149), A Project Funded by the Priority Academic Program Development of Jiangsu Higher Education Institutions (PAPD), The Research Fund of State Key Laboratory of Mechanics and Control of Mechanical Structures (MCMS-I-19G01), and the China Scholarship Council (CSC No. 201906830018).

## Appendix A. Lévy processes

An $n$-dimensional Lévy process $\mathbf{L}_t$ is a stochastic process with the following conditions:

(i) $\mathbf{L}_0 = 0$, a.s.;

(ii) Independent increments: for any choice of $n \geq 1$ and $t_0 < t_1 < \cdots < t_{n-1} < t_n$, the random variables $\mathbf{L}_{t_0}$, $\mathbf{L}_{t_1} - \mathbf{L}_{t_0}$, $\mathbf{L}_{t_2} - \mathbf{L}_{t_1}$, $\cdots$, $\mathbf{L}_{t_n} - \mathbf{L}_{t_{n-1}}$ are independent;

(iii) Stationary increments: $\mathbf{L}_t - \mathbf{L}_s$ and $\mathbf{L}_{t-s}$ have the same distribution;

(iv) Stochastically continuous sample paths: for every $s > 0$, $\mathbf{L}_t \to \mathbf{L}_s$ in probability, as $t \to s$.

The characteristic function for a pure jump $n$-dimensional Lévy process is given by

$$\mathrm{E}\exp(i\mathbf{u}^T \mathbf{L}_t) = \exp\left[ t \int_{\mathbb{R}^n \setminus \{0\}} \left( \mathrm{e}^{i\mathbf{u}^T \mathbf{y}} - 1 - i\mathbf{u}^T \mathbf{y} \chi_{|\mathbf{y}| \leq 1} \right) \nu(\mathrm{d}\mathbf{y}) \right],$$

where the $\chi_S$ is the indicator function of the set $S$. The Lévy jump measure $\nu$ defined on $\mathbb{R}^n \setminus \{0\}$ satisfies the condition

$$\int_{\mathbb{R}^n \setminus \{0\}} \left( |\mathbf{y}|^2 \wedge 1 \right) \nu(\mathrm{d}\mathbf{y}) < \infty,$$

where the sign "$\wedge$" indicates to select the smaller one of the two numbers as the value of the expression. That is, $a \wedge b = \min\{a, b\}$. The Lévy jump measure in fact quantifies the jump frequency and size for sample paths of this Lévy process.

An $\alpha$-stable Lévy motion is a special but most popular type of the Lévy process defined by the stable random variable with the distribution $S_\alpha(\delta, \beta, \lambda)$. Usually, $\alpha \in (0, 2]$ is called the stability parameter, $\delta \in (0, \infty)$ is the scaling parameter, $\beta \in [-1, 1]$ is the skewness parameter and $\lambda \in (-\infty, \infty)$ is the shift parameter.

A stable random variable $X$ with $0 < \alpha < 2$ has the following "heavy tail" estimate:

$$\lim_{y \to \infty} y^\alpha \mathbb{P}(X > y) = C_\alpha \frac{1+\beta}{2} \delta^\alpha;$$

where $C_\alpha$ is a positive constant depending on $\alpha$. In other words, the tail estimate decays polynomially.

In particular, for a rotationally symmetric $\alpha$-stable Lévy process, the characteristic function is

transformed into

$$\mathbb{E}\exp(i\mathbf{u}^T \mathbf{L}_t) = \exp(-t|\mathbf{u}|^\alpha),$$

where the jump measure is given by

$$\nu(d\mathbf{y}) = c(n,\alpha)|\mathbf{y}|^{-(n+\alpha)} d\mathbf{y}$$

with the constant $c(n,\alpha) = \dfrac{\alpha \Gamma((n+\alpha)/2)}{2^{1-\alpha} \pi^{n/2} \Gamma(1-\alpha/2)}$. The $\alpha$-stable Lévy motion has larger jumps with lower jump frequencies for smaller $\alpha$ ($0 < \alpha < 1$), while it has smaller jump sizes with higher jump frequencies for larger $\alpha$ ($1 < \alpha < 2$). The special case $\alpha = 2$ corresponds to (Gaussian) Brownian motion. For more information about Lévy process, refer to Refs. [21,42].

## Appendix B. Stochastic differential equations with Brownian motion and Lévy process

Consider an $n$-dimensional stochastic differential equation

$$d\mathbf{x}(t) = \mathbf{b}(\mathbf{x}(t))dt + \Lambda(\mathbf{x}(t))d\mathbf{B}_t + \sigma d\mathbf{L}_t, \tag{B1}$$

where $\mathbf{b}(\mathbf{x}) = [b_1(\mathbf{x}), \cdots, b_n(\mathbf{x})]^T$ is the drift coefficient in $\mathbb{R}^n$, $\mathbf{B}_t = [B_{1,t}, \cdots, B_{n,t}]^T$ is $n$-dimensional Brownian motion, $\Lambda(\mathbf{x})$ is an $n \times n$ matrix and $\mathbf{L}_t = [L_{1,t}, \cdots, L_{n,t}]^T$ is a symmetric Lévy motion with constant noise intensity $\sigma$. Assume that the initial condition is $\mathbf{x}(0) = \mathbf{z}$, $a(\mathbf{x}) = \Lambda\Lambda^T$ is the diffusion matrix, and the jump measure of $\mathbf{L}_t$ satisfies $\nu(d\mathbf{y}) = W(\mathbf{y})d\mathbf{y}$ for $\mathbf{y} \in \mathbb{R}^n \setminus \{0\}$. Due to the symmetry of the Lévy motion, $W(-\mathbf{y}) = W(\mathbf{y})$.

The generator $A$ for the stochastic differential equation (B1) is

$$Af = \mathbf{b} \cdot \nabla f + \frac{1}{2} Tr[aH(f)] + \text{P.V.} \int_{\mathbb{R}^n \setminus \{0\}} [f(\mathbf{x} + \sigma \mathbf{y}) - f(\mathbf{x})] W(\mathbf{y}) d\mathbf{y}, \tag{B2}$$

where $H(f)$ denotes the Hessian matrix of the function $f$ and P.V. indicates Cauchy Principle Value integral.

**Assumption 1.** (Local Lipschitz condition) For all $R > 0$ and $\mathbf{x}_1, \mathbf{x}_2 \in \mathbb{R}^n$ with $\min\{|\mathbf{x}_1|, |\mathbf{x}_2|\} \leq R$, there exists a constant $K_1 > 0$ such that

$$|\mathbf{b}(\mathbf{x}_1) - \mathbf{b}(\mathbf{x}_2)|^2 + |\Lambda(\mathbf{x}_1) - \Lambda(\mathbf{x}_2)|^2 \leq K_1 |\mathbf{x}_1 - \mathbf{x}_2|^2. \tag{B3}$$

**Assumption 2.** (Linear growth condition) For all $\mathbf{x} \in \mathbb{R}^n$, there exists a constant $K_2 > 0$ such that

$$2\mathbf{x} \cdot \mathbf{b}(\mathbf{x}) + |\Lambda(\mathbf{x})|^2 \leq K_2 (1 + |\mathbf{x}|^2). \tag{B4}$$

Under Assumptions 1 and 2 on the functions $\mathbf{b}$ and $\Lambda$, the stochastic differential equation (B1) has a unique global solution. This solution process is adapted and right-continuous with left limit. This fact has been guaranteed by Refs. [43,44] and Theorem 4.19 in [45]. This solution process can be referred to as a jump-diffusion process.

Furthermore, for the existence and regularity of the probability density for the solution of stochastic differential equation (B1), see Chapter 17-18 of Nuuno et al. [46], and for the nonlocal Fokker-Planck equation satisfied by the probability density, see Duan [21] and Sun et al. [47–49].

## Appendix C. A Lemma about jump measure

**Lemma 1**. *For every* $\varepsilon > 0$ *and* $i, j = 1, 2, \ldots, n$, $\left| \int_{|\mathbf{y}|<\varepsilon} y_i y_j W(\mathbf{y}) \mathrm{d}\mathbf{y} \right| < \infty$.

**Proof**.

According to the appendix, the jump measure satisfies the condition

$$\int_{R^n \setminus \{0\}} (|\mathbf{y}|^2 \wedge 1) \nu(\mathrm{d}\mathbf{y}) < \infty.$$

Then for $\varepsilon \leq 1$,

$$\int_{|\mathbf{y}|<\varepsilon} |\mathbf{y}|^2 W(\mathbf{y}) \mathrm{d}\mathbf{y} \leq \int_{R^n \setminus \{0\}} (|\mathbf{y}|^2 \wedge 1) \nu(\mathrm{d}\mathbf{y}) < \infty.$$

For $\varepsilon > 1$,

$$\int_{1<|\mathbf{y}|<\varepsilon} |\mathbf{y}|^2 W(\mathbf{y}) \mathrm{d}\mathbf{y} \leq \varepsilon^2 \int_{1<|\mathbf{y}|<\varepsilon} W(\mathbf{y}) \mathrm{d}\mathbf{y}$$

$$\leq \varepsilon^2 \int_{R^n \setminus \{0\}} (|\mathbf{y}|^2 \wedge 1) \nu(\mathrm{d}\mathbf{y})$$

$$< \infty.$$

Thus $\int_{|\mathbf{y}|<\varepsilon} |\mathbf{y}|^2 W(\mathbf{y}) \mathrm{d}\mathbf{y} < \infty$. Hence,

$$\left| \int_{|\mathbf{y}|<\varepsilon} y_i y_j W(\mathbf{y}) \mathrm{d}\mathbf{y} \right| \leq \int_{|\mathbf{y}|<\varepsilon} |y_i y_j| W(\mathbf{y}) \mathrm{d}\mathbf{y}$$

$$\leq \int_{|\mathbf{y}|<\varepsilon} |\mathbf{y}|^2 W(\mathbf{y}) \mathrm{d}\mathbf{y}$$

$$< \infty.$$

The proof is complete.

## Appendix D. Proof of Theorem 1

**Proof.**

i) Since $p(\mathbf{x},0|\mathbf{z},0) = \delta(\mathbf{x}-\mathbf{z})$, $p(\mathbf{x},0|\mathbf{z},0) = 0$ for arbitrary $\mathbf{x}$ and $\mathbf{z}$ satisfying $|\mathbf{x}-\mathbf{z}| > \varepsilon$.

Then

$$\lim_{t \to 0} p(\mathbf{x},t|\mathbf{z},0)/t$$
$$= \lim_{t \to 0} \left[ p(\mathbf{x},t|\mathbf{z},0) - p(\mathbf{x},0|\mathbf{z},0) \right]/t$$
$$= \left. \frac{\partial p(\mathbf{x},t|\mathbf{z},0)}{\partial t} \right|_{t=0}$$
$$= -\nabla \cdot \left[ \mathbf{b} p(\mathbf{x},0|\mathbf{z},0) \right] + \frac{1}{2} Tr \left[ H\left( ap(\mathbf{x},0|\mathbf{z},0) \right) \right]$$
$$+ \int_{\mathbb{R}^n \setminus \{0\}} \left[ p(\mathbf{x}+\sigma\mathbf{y},0|\mathbf{z},0) - p(\mathbf{x},0|\mathbf{z},0) \right] W(\mathbf{y}) d\mathbf{y}.$$

Since $p(\mathbf{x},0|\mathbf{z},0) = 0$, $p(\mathbf{x}+\sigma\mathbf{y},0|\mathbf{z},0) = \sigma^{-n}\delta(\mathbf{x}+\sigma\mathbf{y}-\mathbf{z})$ and the symmetry of the function $W$,

the expression $\lim_{t \to 0} p(\mathbf{x},t|\mathbf{z},0)/t$ is reduced to $\sigma^{-n} W\left( \sigma^{-1}(\mathbf{x}-\mathbf{z}) \right)$.

ii) According to the Fokker-Planck equation (2), we have

$$\lim_{t \to 0} t^{-1} \int_{|\mathbf{x}-\mathbf{z}|<\varepsilon} (x_i - z_i) p(\mathbf{x},t|\mathbf{z},0) d\mathbf{x}$$
$$= \lim_{t \to 0} t^{-1} \int_{|\mathbf{x}-\mathbf{z}|<\varepsilon} (x_i - z_i) \left[ p(\mathbf{x},t|\mathbf{z},0) - p(\mathbf{x},0|\mathbf{z},0) + p(\mathbf{x},0|\mathbf{z},0) \right] d\mathbf{x}$$
$$= \int_{|\mathbf{x}-\mathbf{z}|<\varepsilon} (x_i - z_i) \left. \frac{\partial p(\mathbf{x},t|\mathbf{z},0)}{\partial t} \right|_{t=0} d\mathbf{x} + \lim_{t \to 0} \left[ t^{-1} \int_{|\mathbf{x}-\mathbf{z}|<\varepsilon} (x_i - z_i) \delta(\mathbf{x}-\mathbf{z}) d\mathbf{x} \right]$$
$$= -\sum_{j=1}^{n} \int_{|\mathbf{x}-\mathbf{z}|<\varepsilon} (x_i - z_i) \frac{\partial}{\partial x_j} \left[ b_j(\mathbf{x}) p(\mathbf{x},0|\mathbf{z},0) \right] d\mathbf{x}$$
$$+ \frac{1}{2} \sum_{k,l=1}^{n} \int_{|\mathbf{x}-\mathbf{z}|<\varepsilon} (x_i - z_i) \frac{\partial^2}{\partial x_k \partial x_l} \left[ a_{kl}(\mathbf{x}) p(\mathbf{x},0|\mathbf{z},0) \right] d\mathbf{x}$$
$$+ \int_{|\mathbf{x}-\mathbf{z}|<\varepsilon} \int_{\mathbb{R}^n \setminus \{0\}} (x_i - z_i) \left[ p(\mathbf{x}+\sigma\mathbf{y},0|\mathbf{z},0) - p(\mathbf{x},0|\mathbf{z},0) \right] W(\mathbf{y}) d\mathbf{y} d\mathbf{x}.$$

The application of integration by parts into the first term leads to

$$\sum_{j=1}^{n} \int_{|\mathbf{x}-\mathbf{z}|<\varepsilon} (x_i - z_i) \frac{\partial}{\partial x_j} \left[ b_j(\mathbf{x}) p(\mathbf{x},0|\mathbf{z},0) \right] d\mathbf{x}$$
$$= -\sum_{j=1}^{n} \int_{|\mathbf{x}-\mathbf{z}|<\varepsilon} b_j(\mathbf{x}) p(\mathbf{x},0|\mathbf{z},0) \frac{\partial}{\partial x_j} (x_i - z_i) d\mathbf{x}$$
$$= -\sum_{j=1}^{n} \int_{|\mathbf{x}-\mathbf{z}|<\varepsilon} b_j(\mathbf{x}) \delta(\mathbf{x}-\mathbf{z}) \delta_{ij} d\mathbf{x}$$
$$= -b_i(\mathbf{z}).$$

Therein, the boundary condition vanishes since $p(\mathbf{x},0|\mathbf{z},0) = 0$ as $|\mathbf{x}-\mathbf{z}| = \varepsilon$. For the second integration, we use integration by parts twice

$$\sum_{k,l=1}^{n} \int_{|\mathbf{x}-\mathbf{z}|<\varepsilon} (x_i - z_i) \frac{\partial^2}{\partial x_k \partial x_l} [a_{kl}(\mathbf{x}) p(\mathbf{x},0|\mathbf{z},0)] d\mathbf{x}$$

$$= -\sum_{k,l=1}^{n} \int_{|\mathbf{x}-\mathbf{z}|<\varepsilon} \frac{\partial}{\partial x_k} (x_i - z_i) \frac{\partial}{\partial x_l} [a_{kl}(\mathbf{x}) p(\mathbf{x},0|\mathbf{z},0)] d\mathbf{x}$$

$$= \sum_{k,l=1}^{n} \int_{|\mathbf{x}-\mathbf{z}|<\varepsilon} (x_i - z_i) a_{kl}(\mathbf{x}) p(\mathbf{x},0|\mathbf{z},0) \frac{\partial \delta_{ik}}{\partial x_l} d\mathbf{x}$$

$$= 0.$$

For the third integration, we derive it separately. On one hand,

$$\int_{|\mathbf{x}-\mathbf{z}|<\varepsilon} \int_{\mathbb{R}^n \setminus \{0\}} (x_i - z_i) p(\mathbf{x}+\sigma\mathbf{y},0|\mathbf{z},0) W(\mathbf{y}) d\mathbf{y} d\mathbf{x}$$

$$= \int_{|\mathbf{x}-\mathbf{z}|<\varepsilon} \int_{\mathbb{R}^n \setminus \{0\}} (x_i - z_i) \sigma^{-n} \delta(\mathbf{x}+\sigma\mathbf{y}-\mathbf{z}) W(\mathbf{y}) d\mathbf{y} d\mathbf{x}$$

$$= \sigma^{-n} \int_{|\mathbf{x}-\mathbf{z}|<\varepsilon} (x_i - z_i) W(\sigma^{-1}(\mathbf{x}-\mathbf{z})) d\mathbf{x}$$

$$= \sigma^{-n} \int_{|\mathbf{u}|<\varepsilon} u_i W(\sigma^{-1}\mathbf{u}) d\mathbf{u}.$$

Since $W(-\mathbf{y}) = W(\mathbf{y})$, this integration is equal to zero. On the other hand, according to Tonelli's theorem [50], we obtain

$$\int_{|\mathbf{x}-\mathbf{z}|<\varepsilon} \int_{\mathbb{R}^n \setminus \{0\}} (x_i - z_i) p(\mathbf{x},0|\mathbf{z},0) W(\mathbf{y}) d\mathbf{y} d\mathbf{x}$$

$$= \int_{\mathbb{R}^n \setminus \{0\}} \int_{|\mathbf{x}-\mathbf{z}|<\varepsilon} (x_i - z_i) \delta(\mathbf{x}-\mathbf{z}) d\mathbf{x} W(\mathbf{y}) d\mathbf{y}$$

$$= 0.$$

Hence,

$$\lim_{t \to 0} t^{-1} \int_{|\mathbf{x}-\mathbf{z}|<\varepsilon} (x_i - z_i) p(\mathbf{x},t|\mathbf{z},0) d\mathbf{x} = b_i(\mathbf{z}).$$

iii) According to the Fokker-Planck equation (2), we have

$$\lim_{t \to 0} t^{-1} \int_{|\mathbf{x}-\mathbf{z}|<\varepsilon} (x_i - z_i)(x_j - z_j) p(\mathbf{x},t|\mathbf{z},0) d\mathbf{x}$$

$$= \lim_{t \to 0} t^{-1} \int_{|\mathbf{x}-\mathbf{z}|<\varepsilon} (x_i - z_i)(x_j - z_j) [p(\mathbf{x},t|\mathbf{z},0) - p(\mathbf{x},0|\mathbf{z},0) + p(\mathbf{x},0|\mathbf{z},0)] d\mathbf{x}$$

$$= \int_{|\mathbf{x}-\mathbf{z}|<\varepsilon} (x_i - z_i)(x_j - z_j) \frac{\partial p(\mathbf{x},t|\mathbf{z},0)}{\partial t} \bigg|_{t=0} d\mathbf{x}$$

$$+ \lim_{t \to 0} \left[ t^{-1} \int_{|\mathbf{x}-\mathbf{z}|<\varepsilon} (x_i - z_i)(x_j - z_j) \delta(\mathbf{x}-\mathbf{z}) d\mathbf{x} \right]$$

$$= -\sum_{k=1}^{n} \int_{|\mathbf{x}-\mathbf{z}|<\varepsilon} (x_i - z_i)(x_j - z_j) \frac{\partial}{\partial x_k} [b_k(\mathbf{x}) p(\mathbf{x},0|\mathbf{z},0)] d\mathbf{x}$$

$$+ \frac{1}{2} \sum_{k,l=1}^{n} \int_{|\mathbf{x}-\mathbf{z}|<\varepsilon} (x_i - z_i)(x_j - z_j) \frac{\partial^2}{\partial x_k \partial x_l} [a_{kl}(\mathbf{x}) p(\mathbf{x},0|\mathbf{z},0)] d\mathbf{x}$$

$$+ \int_{|\mathbf{x}-\mathbf{z}|<\varepsilon} \int_{\mathbb{R}^n \setminus \{0\}} (x_i - z_i)(x_j - z_j) [p(\mathbf{x}+\sigma\mathbf{y},0|\mathbf{z},0) - p(\mathbf{x},0|\mathbf{z},0)] W(\mathbf{y}) d\mathbf{y} d\mathbf{x}.$$

The application of integration by parts into the first term yields

$$\sum_{k=1}^{n}\int_{|\mathbf{x}-\mathbf{z}|<\varepsilon}(x_i-z_i)(x_j-z_j)\frac{\partial}{\partial x_k}\big[b_k(\mathbf{x})p(\mathbf{x},0\,|\,\mathbf{z},0)\big]d\mathbf{x}$$

$$=-\sum_{k=1}^{n}\int_{|\mathbf{x}-\mathbf{z}|<\varepsilon}b_k(\mathbf{x})p(\mathbf{x},0\,|\,\mathbf{z},0)\frac{\partial}{\partial x_k}\big[(x_i-z_i)(x_j-z_j)\big]d\mathbf{x}$$

$$=-\sum_{k=1}^{n}\int_{|\mathbf{x}-\mathbf{z}|<\varepsilon}b_k(\mathbf{x})\delta(\mathbf{x}-\mathbf{z})\big[\delta_{ik}(x_j-z_j)+\delta_{jk}(x_i-z_i)\big]d\mathbf{x}$$

$$=0.$$

Therein, the boundary condition vanishes since $p(\mathbf{x},0\,|\,\mathbf{z},0)=0$ as $|\mathbf{x}-\mathbf{z}|=\varepsilon$. For the second integration, we use integration by parts again

$$\frac{1}{2}\sum_{k,l=1}^{n}\int_{|\mathbf{x}-\mathbf{z}|<\varepsilon}(x_i-z_i)(x_j-z_j)\frac{\partial^2}{\partial x_k\partial x_l}\big[a_{kl}(\mathbf{x})p(\mathbf{x},0\,|\,\mathbf{z},0)\big]d\mathbf{x}$$

$$=-\frac{1}{2}\sum_{k,l=1}^{n}\int_{|\mathbf{x}-\mathbf{z}|<\varepsilon}\frac{\partial}{\partial x_k}\big[(x_i-z_i)(x_j-z_j)\big]\frac{\partial}{\partial x_l}\big[a_{kl}(\mathbf{x})p(\mathbf{x},0\,|\,\mathbf{z},0)\big]d\mathbf{x}$$

$$=\frac{1}{2}\sum_{k,l=1}^{n}\int_{|\mathbf{x}-\mathbf{z}|<\varepsilon}a_{kl}(\mathbf{x})p(\mathbf{x},0\,|\,\mathbf{z},0)\frac{\partial}{\partial x_l}\big[\delta_{ik}(x_j-z_j)+\delta_{jk}(x_i-z_i)\big]d\mathbf{x}$$

$$=\frac{1}{2}\sum_{k,l=1}^{n}\int_{|\mathbf{x}-\mathbf{z}|<\varepsilon}a_{kl}(\mathbf{x})\delta(\mathbf{x}-\mathbf{z})(\delta_{ik}\delta_{jl}+\delta_{il}\delta_{jk})d\mathbf{x}$$

$$=\frac{1}{2}\big[a_{ij}(\mathbf{z})+a_{ji}(\mathbf{z})\big]$$

$$=a_{ij}(\mathbf{z}).$$

We still derive the third integration separately. On one hand,

$$\int_{|\mathbf{x}-\mathbf{z}|<\varepsilon}\int_{\mathbb{R}^n\setminus\{0\}}(x_i-z_i)(x_j-z_j)p(\mathbf{x}+\sigma\mathbf{y},0\,|\,\mathbf{z},0)W(\mathbf{y})d\mathbf{y}d\mathbf{x}$$

$$=\int_{|\mathbf{x}-\mathbf{z}|<\varepsilon}\int_{\mathbb{R}^n\setminus\{0\}}(x_i-z_i)(x_j-z_j)\delta(\mathbf{x}+\sigma\mathbf{y}-\mathbf{z})W(\mathbf{y})d\mathbf{y}d\mathbf{x}$$

$$=\sigma^{-n}\int_{|\mathbf{x}-\mathbf{z}|<\varepsilon}(x_i-z_i)(x_j-z_j)W\big(\sigma^{-1}(\mathbf{x}-\mathbf{z})\big)d\mathbf{x}$$

$$=\sigma^{-n}\int_{|\mathbf{y}|<\varepsilon}y_iy_jW\big(\sigma^{-1}\mathbf{y}\big)d\mathbf{y}.$$

According to the Lemma in Appendix C, this integration is bounded. On the other hand, according to Tonelli's theorem, we obtain

$$\int_{|\mathbf{x}-\mathbf{z}|<\varepsilon}\int_{\mathbb{R}^n\setminus\{0\}}(x_i-z_i)(x_j-z_j)p(\mathbf{x},0\,|\,\mathbf{z},0)W(\mathbf{y})d\mathbf{y}d\mathbf{x}$$

$$=\int_{|\mathbf{x}-\mathbf{z}|<\varepsilon}\int_{\mathbb{R}^n\setminus\{0\}}(x_i-z_i)(x_j-z_j)\delta(\mathbf{x}-\mathbf{z})W(\mathbf{y})d\mathbf{y}d\mathbf{x}$$

$$=\int_{\mathbb{R}^n\setminus\{0\}}\int_{|\mathbf{x}-\mathbf{z}|<\varepsilon}(x_i-z_i)(x_j-z_j)\delta(\mathbf{x}-\mathbf{z})d\mathbf{x}W(\mathbf{y})d\mathbf{y}$$

$$=0.$$

Hence,

$$\lim_{t\to 0}t^{-1}\int_{|\mathbf{x}-\mathbf{z}|<\varepsilon}(x_i-z_i)(x_j-z_j)p(\mathbf{x},t\,|\,\mathbf{z},0)d\mathbf{x}=a_{ij}(\mathbf{z})+\sigma^{-n}\int_{|\mathbf{y}|<\varepsilon}y_iy_jW\big(\sigma^{-1}\mathbf{y}\big)d\mathbf{y}.$$

The proof is complete.

## Appendix E. Proof of Corollary 2

**Proof.**

i) This is derived directly by integrating the equation in the first assertion of Theorem 1 on the interval $[\varepsilon, m\varepsilon)$.

ii) Let the set $d\Gamma = [u_1, u_1 + du_1) \times [u_2, u_2 + du_2) \times \cdots \times [u_n, u_n + du_n)$. Then we have

$$\mathbb{P}\{\mathbf{x}(t) \in d\Gamma; |\mathbf{x}(t) - \mathbf{z}| < \varepsilon | \mathbf{x}(0) = \mathbf{z}\}$$
$$= \mathbb{P}\{\mathbf{x}(t) \in d\Gamma | \mathbf{x}(0) = \mathbf{z}; |\mathbf{x}(t) - \mathbf{z}| < \varepsilon\} \cdot \mathbb{P}\{|\mathbf{x}(t) - \mathbf{z}| < \varepsilon | \mathbf{x}(0) = \mathbf{z}\}.$$

Thus

$$\int_{|\mathbf{x}-\mathbf{z}|<\varepsilon} (x_i - z_i) p(\mathbf{x}, t | \mathbf{z}, 0) d\mathbf{x}$$
$$= \int_{|\mathbf{u}-\mathbf{z}|<\varepsilon} (u_i - z_i) \mathbb{P}\{\mathbf{x}(t) \in d\Gamma | \mathbf{x}(0) = \mathbf{z}\}$$
$$= \int_{|\mathbf{u}-\mathbf{z}|<\varepsilon} (u_i - z_i) \mathbb{P}\{\mathbf{x}(t) \in d\Gamma; |\mathbf{x}(t) - \mathbf{z}| < \varepsilon | \mathbf{x}(0) = \mathbf{z}\}$$
$$= \mathbb{P}\{|\mathbf{x}(t) - \mathbf{z}| < \varepsilon | \mathbf{x}(0) = \mathbf{z}\} \cdot \int_{|\mathbf{u}-\mathbf{z}|<\varepsilon} (u_i - z_i) \mathbb{P}\{\mathbf{x}(t) \in d\Gamma | \mathbf{x}(0) = \mathbf{z}; |\mathbf{x}(t) - \mathbf{z}| < \varepsilon\}$$
$$= \mathbb{P}\{|\mathbf{x}(t) - \mathbf{z}| < \varepsilon | \mathbf{x}(0) = \mathbf{z}\} \cdot \mathrm{E}\left[(x_i(t) - z_i) | \mathbf{x}(0) = \mathbf{z}; |\mathbf{x}(t) - \mathbf{z}| < \varepsilon\right].$$

Hence, the conclusion is immediately deduced

$$\lim_{t \to 0} t^{-1} \mathbb{P}\{|\mathbf{x}(t) - \mathbf{z}| < \varepsilon | \mathbf{x}(0) = \mathbf{z}\} \cdot \mathrm{E}\left[(x_i(t) - z_i) | \mathbf{x}(0) = \mathbf{z}; |\mathbf{x}(t) - \mathbf{z}| < \varepsilon\right] = b_i(\mathbf{z}).$$

iii) This proof is similar to the second conclusion.

The proof is complete.

## Appendix F. Data Availability Statement

The data that support the findings of this study are openly available in GitHub [51].

Table 1. Identified Lévy motion for the one-dimensional double-well system

| The parameter $\alpha$ | | Lévy noise intensity $\sigma$ | |
|---|---|---|---|
| True | Learning | True | Learning |
| 0.5 | 0.5106 | 2 | 2.0044 |
| 1 | 0.9987 | 2 | 2.0068 |
| 1.5 | 1.4987 | 2 | 2.0196 |

Table 2. Identified drift term for the one-dimensional double-well system

| Basis | True | Learning | | |
|---|---|---|---|---|
| | | $\alpha=0.5$ | $\alpha=1$ | $\alpha=1.5$ |
| 1 | 0 | 0 | 0 | 0 |
| $x$ | 4 | 3.9669 | 3.9968 | 4.0606 |
| $x^2$ | 0 | 0 | 0 | 0 |
| $x^3$ | -1 | -0.9931 | -0.9972 | -1.0093 |
| $x^4$ | 0 | 0 | 0 | 0 |
| $x^5$ | 0 | 0 | 0 | 0 |
| $x^6$ | 0 | 0 | 0 | 0 |

Table 3. Identified diffusion term for the one-dimensional double-well system

| Basis | True | Learning | | |
|---|---|---|---|---|
| | | $\alpha=0.5$ | $\alpha=1$ | $\alpha=1.5$ |
| 1 | 1 | 0.9533 | 0.9895 | 0.9068 |
| $x$ | 2 | 1.9830 | 1.9809 | 1.9723 |
| $x^2$ | 1 | 1.0193 | 1.0201 | 1.0146 |
| $x^3$ | 0 | 0 | 0 | 0 |
| $x^4$ | 0 | 0 | 0 | 0 |
| $x^5$ | 0 | 0 | 0 | 0 |
| $x^6$ | 0 | 0 | 0 | 0 |

Table 4. The stability parameter $\alpha$ versus $\varepsilon$ and $h$

(a) $h=0.001$

| $\varepsilon$ | $\alpha=0.5$ | $\alpha=1$ | $\alpha=1.5$ |
|---|---|---|---|

| 0.1 | 2.3783 | 2.1455 | 2.2266 |
| 0.2 | 1.6429 | 1.6504 | 1.9617 |
| 0.3 | 0.8759 | 1.1899 | 1.6652 |
| 0.5 | 0.5006 | 1.0058 | 1.5210 |
| 1   | 0.4902 | 0.9978 | 1.4856 |

(b) $h = 0.01$

| $\varepsilon$ | $\alpha = 0.5$ | $\alpha = 1$ | $\alpha = 1.5$ |
|---|---|---|---|
| 0.2 | 1.7735 | 1.7302 | 1.9448 |
| 0.5 | 1.2678 | 1.4987 | 1.9874 |
| 1   | 0.5788 | 1.0709 | 1.6412 |
| 1.5 | 0.5051 | 1.0112 | 1.5445 |
| 2   | 0.4986 | 1.0027 | 1.5031 |

Table 5. Identified Lévy motion for the two-dimensional Maier-Stein system

| The parameter $\alpha$ | | Lévy noise intensity $\sigma$ | |
|---|---|---|---|
| True | Learning | True | Learning |
| 0.5 | 0.5038 | 2 | 1.9806 |
| 1   | 1.0013 | 2 | 2.0025 |
| 1.5 | 1.5036 | 2 | 2.0231 |

Table 6. Identified drift term for the two-dimensional Maier-Stein system

(a) The first component $b_1(\mathbf{x})$

| Basis | True | Learning | | |
|---|---|---|---|---|
|   |   | $\alpha = 0.5$ | $\alpha = 1$ | $\alpha = 1.5$ |
| 1 | 0 | 0 | 0 | 0 |
| $x_1$ | 1 | 1.0135 | 0.9808 | 0.9689 |
| $x_2$ | 0 | 0 | 0 | 0 |
| $x_1^2$ | 0 | 0 | 0 | 0 |
| $x_1 x_2$ | 0 | 0 | 0 | 0 |

| Basis | True | | | |
|---|---|---|---|---|
| | | $\alpha=0.5$ | $\alpha=1$ | $\alpha=1.5$ |
| $x_2^2$ | 0 | 0 | 0 | 0 |
| $x_1^3$ | -1 | -1.0051 | -0.9947 | -0.9922 |
| $x_1^2 x_2$ | 0 | 0 | 0 | 0 |
| $x_1 x_2^2$ | -5 | -5.0021 | -4.9839 | -4.9751 |
| $x_2^3$ | 0 | 0 | 0 | 0 |

(b) The second component $b_2(\mathbf{x})$

| Basis | True | Learning | | |
|---|---|---|---|---|
| | | $\alpha=0.5$ | $\alpha=1$ | $\alpha=1.5$ |
| 1 | 0 | 0 | 0 | 0 |
| $x_1$ | 0 | 0 | 0 | 0 |
| $x_2$ | -1 | -1.0046 | -1.0010 | -1.0008 |
| $x_1^2$ | 0 | 0 | 0 | 0 |
| $x_1 x_2$ | 0 | 0 | 0 | 0 |
| $x_2^2$ | 0 | 0 | 0 | 0 |
| $x_1^3$ | 0 | 0 | 0 | 0 |
| $x_1^2 x_2$ | -1 | -0.9997 | -0.9961 | -0.9908 |
| $x_1 x_2^2$ | 0 | 0 | 0 | 0 |
| $x_2^3$ | 0 | 0 | 0 | 0 |

Table 7. Identified diffusion term for the two-dimensional Maier-Stein system

(a) The first component $a_{11}(\mathbf{x})$

| Basis | True | Learning | | |
|---|---|---|---|---|
| | | $\alpha=0.5$ | $\alpha=1$ | $\alpha=1.5$ |
| 1 | 2 | 1.9514 | 1.9395 | 1.8497 |
| $x_1$ | 0 | 0 | 0 | 0 |
| $x_2$ | 2 | 1.9951 | 1.9940 | 1.9883 |

| Basis | True | Learning | | |
|---|---|---|---|---|
| | | $\alpha=0.5$ | $\alpha=1$ | $\alpha=1.5$ |
| $x_1^2$ | 0 | 0 | 0 | 0 |
| $x_1 x_2$ | 0 | 0 | 0 | 0 |
| $x_2^2$ | 1 | 1.1313 | 1.1289 | 1.1268 |
| $x_1^3$ | 0 | 0 | 0 | 0 |
| $x_1^2 x_2$ | 0 | 0 | 0 | 0 |
| $x_1 x_2^2$ | 0 | 0 | 0 | 0 |
| $x_2^3$ | 0 | 0 | 0 | 0 |

(b) The second component $a_{12}(\mathbf{x})$

| Basis | True | Learning | | |
|---|---|---|---|---|
| | | $\alpha=0.5$ | $\alpha=1$ | $\alpha=1.5$ |
| 1 | 0 | 0 | 0 | 0 |
| $x_1$ | 1 | 0.9976 | 0.9966 | 0.9976 |
| $x_2$ | 0 | 0 | 0 | 0 |
| $x_1^2$ | 0 | 0 | 0 | 0 |
| $x_1 x_2$ | 0 | 0 | 0 | 0 |
| $x_2^2$ | 0 | 0 | 0 | 0 |
| $x_1^3$ | 0 | 0 | 0 | 0 |
| $x_1^2 x_2$ | 0 | 0 | 0 | 0 |
| $x_1 x_2^2$ | 0 | 0 | 0 | 0 |
| $x_2^3$ | 0 | 0 | 0 | 0 |

(c) The third component $a_{22}(\mathbf{x})$

| Basis | True | Learning | | |
|---|---|---|---|---|
| | | $\alpha=0.5$ | $\alpha=1$ | $\alpha=1.5$ |
| 1 | 0 | 0 | 0 | 0 |
| $x_1$ | 0 | 0 | 0 | 0 |
| $x_2$ | 0 | 0 | 0 | 0 |
| $x_1^2$ | 1 | 1.0046 | 0.9976 | 0.9616 |
| $x_1 x_2$ | 0 | 0 | 0 | 0 |
| $x_2^2$ | 0 | 0 | 0 | 0 |
| $x_1^3$ | 0 | 0 | 0 | 0 |
| $x_1^2 x_2$ | 0 | 0 | 0 | 0 |
| $x_1 x_2^2$ | 0 | 0 | 0 | 0 |
| $x_2^3$ | 0 | 0 | 0 | 0 |

Table 8. Identified Lévy motion for the three-dimensional Lorenz system

| The parameter $\alpha$ | | Lévy noise intensity $\sigma$ | |
|---|---|---|---|
| True | Learning | True | Learning |
| 0.5 | 0.5031 | 2 | 1.9916 |
| 1 | 1.0006 | 2 | 2.0123 |
| 1.5 | 1.5145 | 2 | 2.0238 |

Table 9. Identified drift term for the three-dimensional Lorenz system

(a) The first component $b_1(\mathbf{x})$

| Basis | True | Learning | | |
|---|---|---|---|---|
| | | $\alpha = 0.5$ | $\alpha = 1$ | $\alpha = 1.5$ |
| 1 | 0 | 0 | 0 | 0 |
| $x_1$ | -10 | -9.9820 | -9.9716 | -9.9482 |
| $x_2$ | 10 | 9.9839 | 9.9570 | 9.9398 |
| $x_3$ | 0 | 0 | 0 | 0 |
| $x_1^2$ | 0 | 0 | 0 | 0 |
| $x_1 x_2$ | 0 | 0 | 0 | 0 |
| $x_1 x_3$ | 0 | 0 | 0 | 0 |

| Basis | | | | |
|---|---|---|---|---|
| $x_2^2$ | 0 | 0 | 0 | 0 |
| $x_2 x_3$ | 0 | 0 | 0 | 0 |
| $x_3^2$ | 0 | 0 | 0 | 0 |

(b) The second component $b_2(\mathbf{x})$

| Basis | True | Learning | | |
|---|---|---|---|---|
| | | $\alpha=0.5$ | $\alpha=1$ | $\alpha=1.5$ |
| 1 | 0 | 0 | 0 | 0 |
| $x_1$ | 4 | 3.9856 | 3.9841 | 3.9885 |
| $x_2$ | -1 | -0.9962 | -1.0000 | -1.0005 |
| $x_3$ | 0 | 0 | 0 | 0 |
| $x_1^2$ | 0 | 0 | 0 | 0 |
| $x_1 x_2$ | 0 | 0 | 0 | 0 |
| $x_1 x_3$ | -1 | -1.0017 | -1.0029 | -0.9994 |
| $x_2^2$ | 0 | 0 | 0 | 0 |
| $x_2 x_3$ | 0 | 0 | 0 | 0 |
| $x_3^2$ | 0 | 0 | 0 | 0 |

(c) The third component $b_3(\mathbf{x})$

| Basis | True | Learning | | |
|---|---|---|---|---|
| | | $\alpha=0.5$ | $\alpha=1$ | $\alpha=1.5$ |
| 1 | 0 | 0 | 0 | 0 |
| $x_1$ | 0 | 0 | 0 | 0 |
| $x_2$ | 0 | 0 | 0 | 0 |
| $x_3$ | -2.6667 | -2.6633 | -2.6577 | -2.6521 |
| $x_1^2$ | 0 | 0 | 0 | 0 |
| $x_1 x_2$ | 1 | 0.9960 | 0.9997 | 1.0006 |
| $x_1 x_3$ | 0 | 0 | 0 | 0 |

| Basis | | | |
|---|---|---|---|
| $x_2^2$ | 0 | 0 | 0 | 0 |
| $x_2 x_3$ | 0 | 0 | 0 | 0 |
| $x_3^2$ | 0 | 0 | 0 | 0 |

Table 10. Identified diffusion term for the three-dimensional Lorenz system

(a) The first component $a_{11}(\mathbf{x})$

| Basis | True | Learning | | |
|---|---|---|---|---|
| | | $\alpha = 0.5$ | $\alpha = 1$ | $\alpha = 1.5$ |
| 1 | 2 | 2.0092 | 2.2601 | 2.0997 |
| $x_1$ | 0 | 0 | 0 | 0 |
| $x_2$ | 0 | 0 | 0 | 0 |
| $x_3$ | 2 | 1.9961 | 1.9893 | 1.9863 |
| $x_1^2$ | 0 | 0.0998 | 0 | 0 |
| $x_1 x_2$ | 0 | -0.1996 | 0 | 0 |
| $x_1 x_3$ | 0 | 0 | 0 | 0 |
| $x_2^2$ | 0 | 0.0996 | 0 | 0 |
| $x_2 x_3$ | 0 | 0 | 0 | 0 |
| $x_3^2$ | 1 | 0.9977 | 0.9935 | 0.9927 |

(b) The second component $a_{12}(\mathbf{x})$

| Basis | True | Learning | | |
|---|---|---|---|---|
| | | $\alpha = 0.5$ | $\alpha = 1$ | $\alpha = 1.5$ |
| 1 | 0 | 0 | 0 | 0 |
| $x_1$ | 0 | 0 | 0 | 0 |
| $x_2$ | 1 | 0.9977 | 0.9954 | 0.9935 |
| $x_3$ | 0 | 0 | 0 | 0 |
| $x_1^2$ | 0 | 0 | 0 | 0 |
| $x_1 x_2$ | 0 | 0 | 0 | 0 |

| Basis | | | | |
|---|---|---|---|---|
| $x_1 x_3$ | 0 | 0 | 0 | 0 |
| $x_2^2$ | 0 | 0 | 0 | 0 |
| $x_2 x_3$ | 0 | 0 | 0 | 0 |
| $x_3^2$ | 0 | 0 | 0 | 0 |

(c) The third component $a_{13}(\mathbf{x})$

| Basis | True | Learning | | |
|---|---|---|---|---|
| | | $\alpha = 0.5$ | $\alpha = 1$ | $\alpha = 1.5$ |
| 1 | 0 | 0 | 0 | 0 |
| $x_1$ | 0 | 0 | 0 | 0 |
| $x_2$ | 0 | 0 | 0 | 0 |
| $x_3$ | 0 | 0 | 0 | 0 |
| $x_1^2$ | 0 | 0 | 0 | 0 |
| $x_1 x_2$ | 0 | 0 | 0 | 0 |
| $x_1 x_3$ | 0 | 0 | 0 | 0 |
| $x_2^2$ | 0 | 0 | 0 | 0 |
| $x_2 x_3$ | 0 | 0 | 0 | 0 |
| $x_3^2$ | 0 | 0 | 0 | 0 |

(d) The fourth component $a_{22}(\mathbf{x})$

| Basis | True | Learning | | |
|---|---|---|---|---|
| | | $\alpha = 0.5$ | $\alpha = 1$ | $\alpha = 1.5$ |
| 1 | 0 | 0 | 0 | 0 |
| $x_1$ | 0 | 0 | 0 | 0 |
| $x_2$ | 0 | 0 | 0 | 0 |
| $x_3$ | 0 | 0 | 0 | 0 |
| $x_1^2$ | 0 | 0 | 0 | 0 |
| $x_1 x_2$ | 0 | 0 | 0 | 0 |

| Basis | | | | |
|---|---|---|---|---|
| $x_1x_3$ | 0 | 0 | 0 | 0 |
| $x_2^2$ | 1 | 1.0127 | 1.0060 | 0.9385 |
| $x_2x_3$ | 0 | 0 | 0 | 0 |
| $x_3^2$ | 0 | 0 | 0 | 0 |

(e) The fifth component $a_{23}(\mathbf{x})$

| Basis | True | Learning | | |
|---|---|---|---|---|
| | | $\alpha=0.5$ | $\alpha=1$ | $\alpha=1.5$ |
| 1 | 0 | 0 | 0 | 0 |
| $x_1$ | 0 | 0 | 0 | 0 |
| $x_2$ | 0 | 0 | 0 | 0 |
| $x_3$ | 0 | 0 | 0 | 0 |
| $x_1^2$ | 0 | 0 | 0 | 0 |
| $x_1x_2$ | 0 | 0 | 0 | 0 |
| $x_1x_3$ | 0 | 0 | 0 | 0 |
| $x_2^2$ | 0 | 0 | 0 | 0 |
| $x_2x_3$ | 0 | 0 | 0 | 0 |
| $x_3^2$ | 0 | 0 | 0 | 0 |

(f) The sixth component $a_{33}(\mathbf{x})$

| Basis | True | Learning | | |
|---|---|---|---|---|
| | | $\alpha=0.5$ | $\alpha=1$ | $\alpha=1.5$ |
| 1 | 0 | 0 | 0 | 0 |
| $x_1$ | 0 | 0 | 0 | 0 |
| $x_2$ | 0 | 0 | 0 | 0 |
| $x_3$ | 0 | 0 | 0 | 0 |
| $x_1^2$ | 1 | 1.0076 | 1.0028 | 0.9327 |
| $x_1x_2$ | 0 | 0 | 0 | 0 |
| $x_1 x_3$ | 0 | 0 | 0 | 0 |
| $x_2^2$ | 0 | 0 | 0 | 0 |
| $x_2 x_3$ | 0 | 0 | 0 | 0 |
| $x_3^2$ | 0 | 0 | 0 | 0 |

Table 11. Identified Lévy motion for the genetic regulatory system

| The parameter $\alpha$ | | Lévy noise intensity $\sigma$ | |
|---|---|---|---|
| True | Learning | True | Learning |
| 0.5 | 0.4939 | 2 | 2.0674 |
| 1 | 1.0175 | 2 | 1.9651 |
| 1.5 | 1.5199 | 2 | 2.0669 |